\def\year{2022}\relax
\documentclass[letterpaper]{article} 
\usepackage{aaai22}  
\usepackage{times}  
\usepackage{helvet}  
\usepackage{courier}  
\usepackage[hyphens]{url}  
\usepackage{graphicx} 
\usepackage{natbib}  
\usepackage{caption} 
\DeclareCaptionStyle{ruled}{labelfont=normalfont,labelsep=colon,strut=off} 
\usepackage{csquotes}
\usepackage{epigraph} 
\usepackage{cancel}
\usepackage{enumerate}

\usepackage{subcaption}
\usepackage{xspace}
\usepackage{amssymb}
\usepackage{multirow}
\usepackage{soul} 
\usepackage{xcolor}
\usepackage{xspace}
\usepackage{amsmath}

\newcommand\algname[1]{\textsf{#1}\xspace}
\newcommand\cbs{\algname{CBS}}
\newcommand\pbs{\algname{PBS}}
\newcommand\tfcbs{\algname{TF-CBS}}
\newcommand\tfpbs{\algname{TF-PBS}}

\newcommand{\mapf}{MAPF\xspace}
\newcommand{\tmapf}{tMAPF\xspace}
\newcommand{\lmapf}{L-MAPF\xspace}
\newcommand{\mapd}{MAPD\xspace}

\newcommand{\cmapf}{C-MAPF\xspace}

\newcommand{\smallEnv}{\textsc{Small}\xspace}
\newcommand{\largeEnv}{\textsc{Large}\xspace}

\def\A{\mathcal{A}}
\def\O{\mathcal{O}}
\def\G{\mathcal{G}}
\def\V{\mathcal{V}}
\def\E{\mathcal{E}}

\newcommand{\Vstart}{\ensuremath{\V_{\text{start}}}\xspace}
\newcommand{\Vgoal}{\ensuremath{\V_{\text{goal}}}\xspace}

\newcommand{\ignore}[1]{}

\newcommand{\OS}[1]{{\textcolor{orange}{\textbf{OS:} #1}}}

\usepackage[vlined, ruled, linesnumbered]{algorithm2e} 
\usepackage{tikz}
\usetikzlibrary{fit,calc}

\newcommand*{\tikzmk}[1]{
    \tikz[remember picture,overlay,] \node (#1) {};\ignorespaces
}
\newcommand{\boxit}[2]{
    \tikz[remember picture,overlay]{\node[rounded corners=4pt,xshift=#2,yshift=1.75pt,fill=#1,opacity=.25,fit={(A)($(B)+(.9\linewidth,.9\baselineskip)$)}] {};}\ignorespaces
}
\colorlet{highlight}{blue!40} 

\newcommand*\rotate{\rotatebox{90}}

\title{Multi-Agent Terraforming:\\Efficient Multi-Agent Path Finding via Environment Manipulation}

\author{
    David Vainshtein,
    Kiril Solovey, 
    Oren Salzman
}
\affiliations{
    Technion Israel Institute of Technology\\
    Haifa, Israel\\
    dudiwa@campus.technion.ac.il, kirilsol@technion.ac.il, osalzman@cs.technion.ac.il
}

\begin{document}

\maketitle

\begin{abstract}
Multi-agent pathfinding (\mapf) is concerned with planning collision-free paths for a team of agents from their start to goal locations in an environment cluttered with obstacles. Typical approaches for \mapf consider the locations of obstacles as being \emph{fixed}, which limits their effectiveness in automated warehouses, where obstacles (representing pods or shelves) can be \emph{moved} out of the way by agents (representing robots) to relieve bottlenecks and introduce shorter routes. 
In this work we initiate the study of \mapf with movable obstacles. In particular, we introduce a new extension of \mapf, which we call Terraforming \mapf (\tmapf), where some agents are responsible for moving obstacles to clear the way for other agents. Solving \tmapf is extremely challenging as it requires reasoning not only about collisions between agents, but also \emph{where} and \emph{when} obstacles should be moved. 
We present extensions of two state-of-the-art algorithms, \cbs and \pbs, in order to tackle \tmapf, and demonstrate that they can 
consistently outperform the best solution possible under a static-obstacle setting.
\ignore{
A recent work~\cite{vainshtain2021multi} attempted to lift the static-obstacle assumption through the introduction of the Terraforming \mapf (\tmapf) problem, with \emph{self-propelled} (i.e., can move without agent intervention) obstacles, which is unlikely to be true in practice. In this work, we consider a more realistic extension of \tmapf with \emph{movable} obstacles that can be carried by agents.   }
\end{abstract}

\section{Introduction}

\epigraph{The impediment to action advances action. \\
What stands in the way becomes the way.}{Marcus Aurelius}

Multi-agent path finding (\mapf) is a popular algorithmic framework that captures complex tasks involving mobile agents that need to plan individual routes while avoiding collisions during plan execution~\cite{stern2019multi,SalzmanS20}. 
This abstraction has been successfully applied to a variety of settings (see, e.g.,~\citet{wurman2008cooperative,BelovDBHKW20,LCZCHS0K21,Choudhury.ea.21}). However, in some cases this formulation may not be expressive enough to fully capture the underlying task, which can lead to suboptimal performance. 

This is especially true in the context of automated warehouses where we are given a stream of tasks, and the goal is to maximize the system's throughput.
%
In this setting, 
formulated as a \emph{lifelong} \mapf (\lmapf) problem and typically solved via a sequence of  \mapf queries~\cite{ma2017lifelong,liu2019task,li2021lifelong},
inventory pods that hold goods are manipulated by a large team of mobile agents (or robots):  
agents pick up pods, carry them to designated dropoff locations where goods are manually removed from the pods (to be packaged for customers); each pod is then carried back by a robot to a (possibly different) storage location~\cite{wurman2008cooperative}. When applied to this setting, existing variants of \mapf and \lmapf
tend to impose the following limiting and artificial constraint: pods that are not currently carried to or from a dropoff location are modeled as \emph{static obstacles}, which cannot be moved. Thus, those approaches overlook the fact that pods can be manipulated to clear the way for agents and reduce the travel time or distance of agents in the system.

\begin{figure}[tb]
     \centering
     \begin{subfigure}[b]{0.3\columnwidth}
         \centering
         \includegraphics[width=\columnwidth]{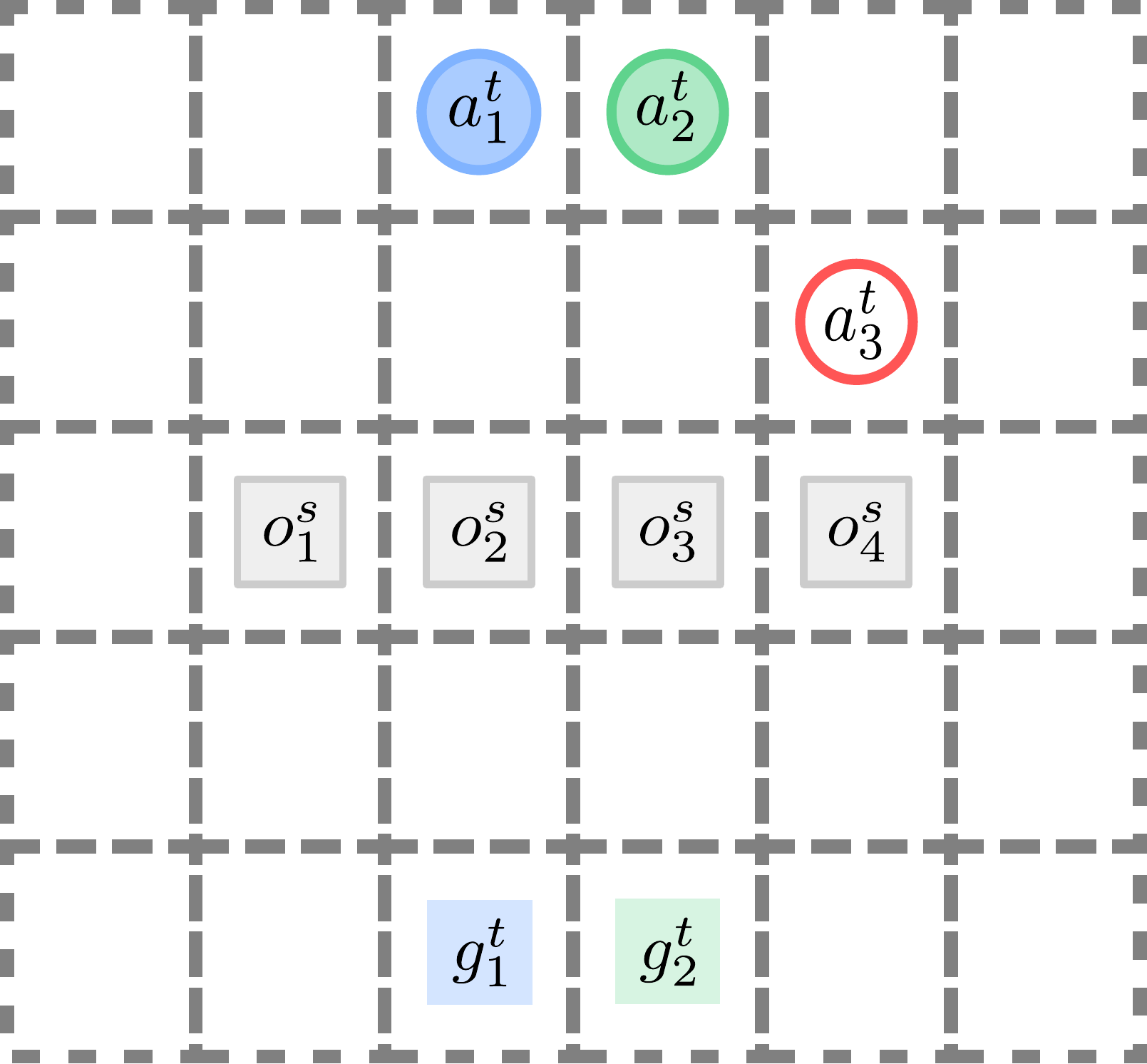}
         \caption{MAPF problem}
         \label{fig:tMAPF1}
     \end{subfigure}
     \hfill
     \begin{subfigure}[b]{0.3\columnwidth}
         \centering
         \includegraphics[width=\columnwidth]{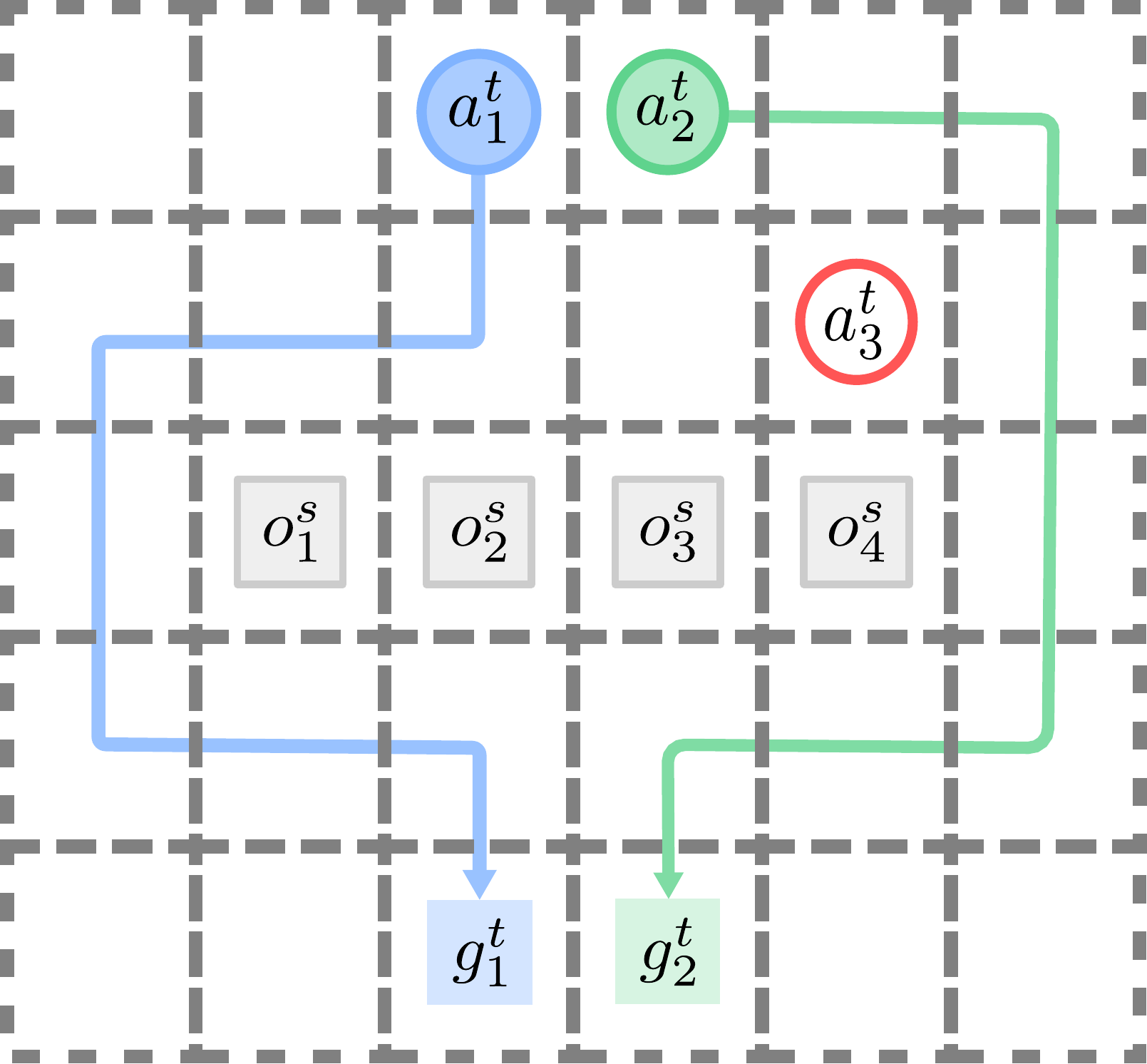}
         \caption{Static obstacles}
         \label{fig:tMAPF2}
     \end{subfigure}
     \hfill
     \begin{subfigure}[b]{0.3\columnwidth}
         \centering
         \includegraphics[width=\columnwidth]{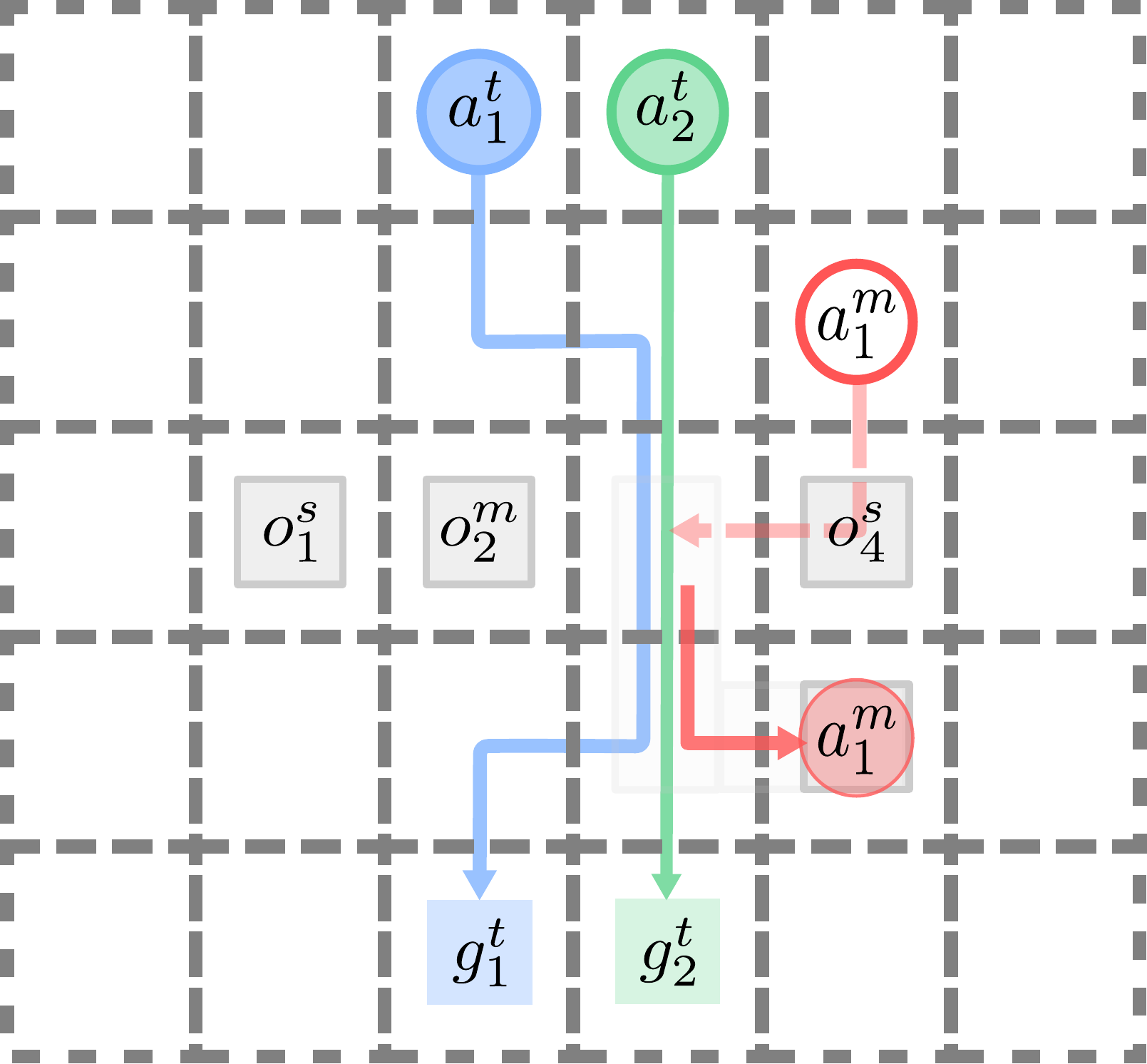}
         \caption{Terraforming}
         \label{fig:tMAPF3}
     \end{subfigure}
    \caption{
        Comparing \mapf and \tmapf for a toy problem with agents $\textcolor{blue}{a^t_1},\textcolor{teal}{a^t_2},\textcolor{red}{a_3}$ (circles), and a row of obstacles (grey squares). 
        (\subref{fig:tMAPF1})~\mapf problem assigning agents to their goal locations (squares). Here, agent $\textcolor{red}{a_3}$ is a task agent (denoted as $\textcolor{red}{a_3^t}$) with no task assigned to him.
        (\subref{fig:tMAPF2})~\mapf solution where agents must avoid collisions with obstacles and each other. Here, agent~$\textcolor{red}{a_3}$ cannot help the other two agents as all obstacles are static.
        (\subref{fig:tMAPF3})~\tmapf solution where $\textcolor{red}{a_3}$ is a mover agent (denoted as $\textcolor{red}{a_1^m}$) that creates a shortcut by clearing a movable obstacle $o_3^m$. Video \textcolor{blue}{\textsf{https://bit.ly/3ImgfAw}}}
        \vspace{-5mm}
    \label{fig:mapf_intro}
\end{figure}

To bridge the gap between existing \mapf formulations and the type of problems they are intended to tackle in the real world, we explore the implications of allowing agents the extra flexibility of \emph{manipulating the environment by moving obstacles} (e.g., dynamically relocating pod's locations in warehouse applications).
To this end, we introduce a new \mapf variant which we term \emph{``Terraforming \mapf''} (or \tmapf in short)\footnote{
The term ``Terraforming'', which originated from SciFi literature and was recently used in the context of space exploration, is the process of deliberately altering the environment of a planet to make it habitable.
}.
In \tmapf, formally defined in Sec.~\ref{sec:problem}, the input consist of  the agents' start and goal location and a set of obstacles (as in \mapf), as well as a specification of the obstacles whose position can be changed while answering the \tmapf query (see Fig.~\ref{fig:mapf_intro} and details below). 

We suggest two algorithms to solve the  \tmapf problem which are described in Section~\ref{sec:algs}.
The first, which we call \tfcbs, is based on the celebrated \cbs algorithm~\cite{sharon2015conflict} which we review in Sec.~\ref{sec:background}.
\tfcbs is complete and is guaranteed to produce a cost-optimal path.
The second algorithm, which we call \tfpbs, 
offers computational efficiency by trading completeness and optimality guarantees in favor of rapidly attaining high-quality solutions. In our evaluation, described in Sec.~\ref{sec:eval}, we demonstrate how both algorithms consistently out-perform the optimal solution produced by classical \mapf algorithms on several different metrics.

In this work we concentrate on the algorithmic implication of terraforming, or environment manipulation, in the context of \mapf.
However, our ultimate goal is to apply terraforming to the \lmapf problem.
As we will see, this is extremely challenging which is why in this work we limit ourselves to a simplified \tmapf setting.
We discuss the research challenges of moving from terraforming in the context of \mapf to \lmapf in Sec.~\ref{sec:discussion}.

\section{Related work}
A variety of approaches were developed to solve the \mapf problem and its many variants using algorithmic tools such as network flow~\cite{YL12}, 
satisfiability~\cite{SFSB16},
Answer Set Programming~\cite{EKOS13}
and search-based methods~\cite{BSSF14,BFSSTBS15,sharon2015conflict,li2019disjoint}.

In this work we adapt approaches from the latter group. Of specific relevance to our work are Conflict-Based Search (\cbs)~\cite{sharon2015conflict} which is used as an algorithmic building block in many state-of-the-art \mapf solvers (see, e.g.~\citet{GGSS21})  
and
Priority-Based Search (\pbs)~\cite{ma2019searching} which is commonly used when solving the \lmapf problem~\cite{ma2017lifelong,li2021lifelong}.
As both are used to develop our new algorithms, we elaborate on both algorithms in Sec.~\ref{sec:background}.

Arguably, the most closely-related work to our new problem formulation is by \citet{bellusci2020multi} who introduce a new variant of the \mapf problem, called Configurable \mapf (\cmapf). 
Here, the structure of the environment is configurable (given a set of constraints) and the problem calls for computing an environment configuration as well as a solution to a \mapf query.
Note that a solution to the  \cmapf problem is  a \emph{fixed} environment (as well as a set of paths dictating where each agent should go).
In contrast, in \tmapf the environment \emph{dynamically changes} as agents execute their paths.

Finally, we mention the works of~\citet{hauser2013minimum,hauser2014minimum} that focus on computing paths that minimize the number of obstacles to remove from a given environment. However, these works are formulated for a single-agent and even then, the agent is not tasked with removing said obstacles.
Also related is the work on Multi-Robot Clutter Removal (MRCR)~\cite{tang2020computing}.
Here, a group of agents are tasked with clearing  obstacles from a given environment while  avoiding collisions with each other and any remaining obstacles.

\section{Problem Formulation}\label{sec:problem}
In this section we provide a formal definition of the \tmapf problem. In preparation, we first formally define the (classical) \mapf problem. 
%

\subsection{Multi-Agent Path Finding (\mapf)}
A \mapf problem is a tuple 
$$
\langle
 \G, \A, \O, \Vstart, \Vgoal
\rangle,
$$ where 
$\G=(\V,\E)$ 
is the environment graph, 
$\A=\{a_1,\ldots,a_n\}$ is the agent set, 
$\O=\{o_1,\ldots,o_\ell\}\subset \V$ is the obstacle set, 
$\Vstart = \{s_1, \ldots, s_n \} \subset \V$ is the set of agents' start vertices
and
$\Vgoal = \{g_1, \ldots, g_n \} \subset \V$ is the set of agents' goal vertices. 
We now elaborate on those ingredients.  

We assume that the graph $\G=(\V,\E)$  on which the agents~$\A$ operate is undirected and includes self edges to all the vertices to simulate agent wait actions, i.e., $(v,v)\in \E$ for every vertex $v\in \V$ and when an agent moves along such an edge we will say that it waits in place.
In typical \mapf formulations, \emph{static obstacles}, which block certain agent positions, are implicitly encoded via the graph $\G$, where vertices representing static obstacle locations are removed. In our setting, in preparation to the \tmapf problem where some agents can share locations with obstacles, it will be convenient to explicitly account for obstacles. In particular, the set of obstacles $\O$ denotes graph vertices that are blocked, and which the agents cannot visit. 

\paragraph{States.}
A \mapf \emph{state} $S=(v_1, \ldots, v_m)$ is a vector where $v_i\in \V$ represents the location of agent $a_i$.
We say that state~$S$ is \emph{valid} if the following conditions are met:
\begin{itemize}
    \item[\textbf{S1}]
    No two agents share the same location, i.e., 
    $v_i\neq v_j$ for any two agents $a_i\neq a_j$.
    \item[\textbf{S2}]
    Agents do not collide with obstacles, i.e.,  $v_i\not \in \O$ for any agent $a_i$.
\end{itemize}

\paragraph{Transitions.}
A \emph{transition} between two valid states  $S=(v_1, \ldots, v_m)$ and $S'=(v'_1, \ldots, v'_m)$ is \emph{valid} if the following conditions are met:
\begin{itemize}
    \item[\textbf{T1}] 
    Agents move along edges, i.e.,
    $\forall a_i\in \A,~(v_i,v'_i)\in \E$.
    \item[\textbf{T2}]  
    Agents do not swap locations over the same edge, i.e.,
    $\forall a_i, a_j \in \A~s.t.~i\neq j$ it holds that $v_i\neq v'_j$ or $v'_i\neq v_j$.
\end{itemize}


\paragraph{Solution.}
A \emph{solution} to the above \mapf problem is a sequence of states $\pi=(S^1,\ldots, S^k)$, such that each state $S^i:=(v_1^i,\ldots, v_n^i)$ is valid, each transition from $S^i$ to $S^{i+1}$ is valid for any 
$i$,
$S^1=\Vstart$
and $S^k=\Vgoal$.
%

\paragraph{Solution cost in \mapf.}
To define the cost of a solution $\pi$ we first define a $\text{cost}(\pi, a_i)$ for agent~$a_i\in \A$ to be the earliest arrival time to its goal after which $a_i$  does not change its location.
Namely, $\text{cost}(\pi, a_i)$ is the smallest $j$ s.t.\ $\forall \tau \in [j,k], v^\tau = g_i$.
The cost of a solution is then defined as 
$\text{Cost}(\pi):= \sum_i \text{cost}(\pi, a_i)$
and is commonly referred to as the \emph{sum of costs} (we explain in Sec.~\ref{subsec:notes} why we use this cost function and not other commonly-used ones such as the makespan~\cite{stern2019multi}).

\subsection{Terraforming \mapf}
A \tmapf problem, which generalizes the \mapf problem (and is visualized in Fig.~\ref{fig:mapf_intro}), is a tuple 
\begin{multline*}
\langle
 \G, 
 \A = \A^t \cup \A^m, 
 \O = \O^s \cup \O^m,\\ 
 \Vstart=\Vstart^t\cup\Vstart^{ma}\cup \Vstart^{mo},
 \Vgoal^t
\rangle.
\end{multline*}
Here, $\G=(\V,\E)$ is a graph defined as in the \mapf setting. 
However, now there are two types of agents:~$n_t$ \emph{task agents}~$\A^t$,
and 
$n_m$ \emph{mover agents} $\A^m$.
Similar to the \mapf setting, task agents $\A^t = \{a_1^t, \ldots, a^t_{n_t} \}$
are forbidden from moving to either of the two types of obstacle vertices (see below) and have designated start vertices $\Vstart^t = \{s_1^t, \ldots, s^t_{n_t} \} \subset \V$
and goal vertices $\Vgoal^t = \{g_1^t, \ldots, g^t_{n_t} \}\subset \V$.
In contrast, mover agents $\A^m = \{a_1^m, \ldots, a^m_{n_m} \}$ can share locations with obstacles (to simulate robots going underneath pods in automated warehouses), and only have designated start vertices 
$\Vstart^{ma} = \{s_1^{ma}, \ldots, s^{ma}_{n_m} \} \subset \V$ without predefined goal vertices.\footnote{As a convention, superscript `t' corresponds to task agents,
superscript `s' corresponds to static obstacles
while superscript `m' corresponds to either mover agents or movable obstacles.
When we need to distinguish between the latter two we use superscripts `ma' and `mo' to refer to mover agents and movable obstacles, respectively.}

In \tmapf we have two types of obstacles:
$\ell_s$ \emph{static obstacles} $\O^s$, 
and 
$\ell_m$ \emph{movable obstacles} $\O^m$ whose location can change via a mover agents (to be explained shortly). Static obstacles $\O^s=\{o_1^s,\ldots,o^s_{\ell_s}\}$ are associated with their vertex locations, i.e., $\O^s\subset \V$, as in the \mapf setting. 
Every movable obstacles $o^m_i\in \O^m$ has a unique start position $s_1^{mo} \in \Vstart^{mo} \subset \V $ that also serves as its goal position. The location of a movable obstacle can change throughout the execution via the mover agents. 

\paragraph{States.}
A \tmapf state encodes the locations of the two types of agents, as well as the locations of the movable obstacles. 
In particular, a state  
$S = (
v_1,\ldots, v_{n_t}, 
u_1,\ldots, u_{n_m}, 
w_1,\ldots,w_{\ell_m})$ 
is a vector of vertices, 
where $v_i\in V $ represents the location of task agent $a^t_i\in \A^t$, 
$u_j\in V$ represents the location of a mover agent $a^m_j\in \A^m$, 
and $w_k\in V$ represents the location of the movable obstacles $o^m_k\in \O^m$. 
A state $S$ is \emph{valid} if the following conditions are met:
\ignore{
(i')~No two agents share the same vertex, i.e., 
$v_i\neq v_j$ for any $a^t_i\neq a^t_j$, 
$u_i\neq u_j$ for any $a^m_i\neq a^m_j$, 
and 
$v_i\neq u_j$ for any $a^t_i\in \A, a^m_j\in \A^m$.
(ii')~No two obstacles share the same vertex, i.e., 
$w_i\neq w_j$ and $w_i,w_j \not \in \O^s$ for any $o^m_i\neq o^m_j \in \O^m$.
(iii')~Task agent do not collide with obstacles, i.e., $v_i\not\in \O^m$ and $v_i\neq w_j$ for any $a_i\in \A$ and $o^m_j\in \O^m$.  
}

\begin{itemize}
    \item[\textbf{S1'}]
    No two agents share the same vertex, i.e., 
    $v_i\neq v_j$ for any $a^t_i\neq a^t_j$, 
    $u_i\neq u_j$ for any $a^m_i\neq a^m_j$, 
    and 
    $v_i\neq u_j$ for any $a^t_i\in \A, a^m_j\in \A^m$.
    \item[\textbf{S2'}] 
    No two obstacles share the same vertex, i.e., 
    $w_i\neq w_j$ and $w_i,w_j \not \in \O^s$ for any
    $o^m_i\neq o^m_j \in \O^m$.
    \item[\textbf{S3'}]
    No task agents collide with obstacles, i.e., $v_i\not\in \O^m$ and $v_i\neq w_j$ for any $a_i\in \A$ and $o^m_j\in \O^m$.
\end{itemize}

\paragraph{Transitions.}
A \emph{transition} between two valid states  
$S=(
v_1,\ldots, v_{n_t}, 
u_1,\ldots, u_{n_m}, 
w_1,\ldots,w_{\ell_m})$ 
and 
$S'=(
v'_1,\ldots, v'_{n_t}, 
u'_1,\ldots, u'_{n_m}, 
w'_1,\ldots,w'_{\ell_m})$ 
is \emph{valid} if the following conditions are met:
\begin{itemize}
    \item[\textbf{T1'}]  
    Agents and movable obstacles move along graph edges, i.e., 
    $\forall i~(S[i],S'[i])\in \E$
    where $S[i]$ ($S'[i]$) denotes the $i$'th element of $S$ ($S'$).
    \item[\textbf{T2'}]  Agents and movable obstacles do not swap locations using the same edge, i.e.,
    $\forall i\neq j~S[i] \neq S'[j]$ or $S[j]\neq S'[i]$.
    \item[\textbf{T3'}] Movable obstacles can only move via a mover agent, i.e.,
    if $w_i \neq w'_i$, for some movable obstacle $o^m_i\in \O^m$, then there exists a mover $a^m_j \in \A^m$ such that $w_i=u_j$ and $w'_i=u'_j$. 
\end{itemize}

\paragraph{Solution.}
A \emph{solution} to the above \tmapf problem  is a sequence of states $\pi=(S^1,\ldots, S^k)$, such that each state 
$S^i:=(v_1^i,\ldots, v_{n_t}^i, u_1^i,\ldots, u_{n_m}^i, w_1^i,\ldots,w_{\ell_m}^i)$ is valid, each transition from $S^i$ to $S^{i+1}$ is valid for any $i$,
$S^1=(s^t_1,\ldots,s^t_{n_t},s^{ma}_1,\ldots, s^{ma}_{n_m},s^{mo}_1,\ldots, s^{mo}_{\ell_m})$, 
$v^k_i=g^t_i$ for any task agent $a^t_i\in \A^t$
and  $w^k_i=s^{mo}_i$ for any movable obstacle $o^m_i\in \O^m$. 

\paragraph{Simplifying assumptions.}
In this work we impose several simplifying assumptions on our \tmapf problem:
\begin{itemize}
    \item[\textbf{A1}]  A mover agent can move at most one movable obstacle. 
    \item[\textbf{A2}] The number of movable obstacles is equal to the number of mover agents.
\end{itemize}
In Sec.~\ref{sec:discussion} we discuss how these simplifying assumptions can be lifted in future work.

\paragraph{Solution cost in \tmapf.}
We consider several cost metrics for a given solution $\pi=(S^1,\ldots, S^k)$.
In preparation, we define different cost functions for the different agents and for the movable obstacles.
Similar to the \mapf setting, for a task agent $a^t_i\in \A^t$, we define 
$\text{cost}^t(\pi, a^t_i)$ to be the smallest~$j$ s.t.~$\forall \tau \in [j,k], v_i^\tau = g_i$. We define the total task-agent cost as $\text{Cost}^t(\pi) = \sum_i \text{cost}^t(\pi, a^t_i)$.

For a mover agent $a^m_i\in \A^m$, denote $\text{cost}^{ma}_{\bcancel{w}}(\pi,a^m_i)$ to be the number of steps to reach the movable obstacle it is going to move without accounting for wait actions, i.e., the cost of a wait action is zero. The total cost for mover agents to reach the movable obstacles is $\text{Cost}^{ma}_{\bcancel{w}}(\pi) = \sum_i \text{cost}^{ma}_{\bcancel{w}}(\pi, a^m_i)$. 
Additionally, define  $\text{cost}^{mo}_{\bcancel{w}}(\pi,a^m_i)$ 
to be the number of steps a movable obstacle takes without wait actions
and define the total cost for movable  obstacles to be
$\text{Cost}^{mo}_{\bcancel{w}}(\pi) = \sum_i \text{cost}^{mo}_{\bcancel{w}}(\pi, a^m_i)$.

In this work we will consider the following two cost functions accounting for motions taken by both types of agents as well as the moveable obstacles.
\begin{equation}
\label{eq:without-mover}
    \text{Cost}_1(\pi) = 
        \text{Cost}^t(\pi)
        +
        \text{Cost}^{mo}_{\bcancel{w}}(\pi).
\end{equation}
\begin{equation}
\label{eq:with-mover}
    \text{Cost}_2(\pi) = 
        \text{Cost}^t(\pi)
        +
        \text{Cost}^{ma}_{\bcancel{w}}(\pi)
        +
        \text{Cost}^{mo}_{\bcancel{w}}(\pi).
\end{equation}

In both cost functions we treat task agents just as in the \mapf setting and the difference arises in how we treat mover agents and movable obstacles.
In addition, in both cost functions we do not account for wait actions (see discussion in Sec.~\ref{subsec:notes}) incurred by the mover agents and movable obstacles. 
Intuitively, in Eq.~\eqref{eq:without-mover} we ignore the cost of reaching a movable obstacle by a mover agent and only account for the ``work'' required to move movable obstacle.
In Eq.~\eqref{eq:with-mover} we add the motions required by a mover agent.

\subsection{Discussion}
\label{subsec:notes}
When formalizing the \tmapf problem, there are many subtle-yet-important variants one can consider. 
E.g., ``what cost function to use?'' and ``are mover agents allowed to move under static obstacles?''.
In our formulation, we made sure to 
have our \tmapf formulation a \emph{generalization} of the standard \mapf formulation
while at the same time
serve as a stepping stone to our ultimate goal of applying terraforming to the \lmapf problem.
Indeed, in Sec.~\ref{sec:discussion} we discuss what are the steps and challenges required to reach this goal.

Specifically, in the \mapf setting we chose to use the sum of costs as our cost function as it extends naturally to the lifelong setting where we wish to maximize a system's throughput.
Moving to the terraforming version of \lmapf in warehouse applications, we envision that there will be no pre-allocation of the agents to two distinct groups of task and mover agents. These two groups will form naturally where agents en-route to a pick an item (obstacle in our formulation) will serve as mover agents while agents already carrying said items will serve as task agents.
This motivated us to
(i)~allow mover agents to move under static obstacles 
and
(ii)~to propose cost functions that focus on the work done to move movable obstacles and not necessarily to reach them as we assume that this will be done by agents en-route to reaching a goal.
Thus, Eq.~\eqref{eq:without-mover} may be seen as a \emph{lower-bound} on the cost to move obstacles as we do not account for the steps taken to reach it and assume that the mover agent would have passed next to the movable obstacle.
Similarly, Eq.~\eqref{eq:with-mover} may be seen as an \emph{upper-bound} on the cost to move obstacles as we assume that at least some of the steps taken to reach it would have been carried out regardless.



\section{Algorithmic Background}\label{sec:background}
Before we present our approaches for \tmapf, we describe in this section two algorithmic building blocks. Namely, the \algname{CBS} and \algname{PBS} algorithms for (classical) \mapf.

\subsection{Conflict-Based Search}
Conflict-Based Search (\cbs) is a popular approach for the (classical) \mapf problem, which is both complete and optimal.
We now provide an overview of \cbs. Specifically, we describe a recent variant by~\citet{li2019disjoint} that uses both positive and negative constraints (to be explained shortly) as it was shown to have better runtime both empirically and when looking at a worst-case complexity analysis~\cite{gordon2021revisiting}.
We refer the reader to~\cite{sharon2015conflict} and~\cite{li2019disjoint}  for the full description.

\cbs maintains constraints between agents which are used to resolve conflicts.
On the high-level search, \cbs explores a \emph{constraint tree} (CT), where a given node~$N$ of CT encodes a set of constraints $C_N$ on the locations of agents in time and space. In particular, \cbs includes positive constraints denoted by $(+,a_i,v,\tau)$ which specify that an agent~$a_i$ must visit vertex $v$ at time step $\tau$, as well as negative constraints denoted by $(-,a_i,v',\tau')$ which prohibits agent $a_i$ from visiting vertex $v'$ at time step $\tau'$. Similar constraints are imposed on edges. In addition to the constraints, each CT node maintains single-agent paths that represent the current \mapf solution (possibly containing conflicts) as well as the cost of the solution. 

\cbs starts the high-level search with the tree root whose constraint set is empty, and assigns to each agent its shortest path, while avoiding static obstacles but ignoring interactions between agents. Whenever \cbs expands a node~$N$, it invokes a low-level search to compute a new set of paths that abide by the constraint set $C_N$ (see details below). If a collision between agents, e.g., $a_i$ and $a_j$, at a vertex~$v$ (or an edge) at time step $\tau$ is encountered in the new  paths, \cbs generates two child CT nodes $N_1, N_2$ with the updated constraints $C_{N_1}=C_N\cup \{(+,a_i,v,\tau)\}, C_{N_2}=C_N\cup \{(-,a_i,v,\tau)\}$, respectively. That is, the node $N_1$ includes constraints to enforce that $a_i$ visits $v$ at time $\tau$ which implicitly adds the constraints $(-,a_\ell,v,\tau)$ for $\ell \neq i$. The node~$N_2$ encodes the opposite situation with respect to $a_i$ forcing it to avoid $v$ at time $\tau$. The high-level search chooses to expand at each iteration a CT node with the lowest cost. The high-level search terminates when a valid solution is found at some node $N$, or when no more nodes for expansion remain, in which case, \cbs declares failure. 

The low-level search of \cbs proceeds as follows: For a given CT node $N$, an \algname{A*}~\cite{HNR68} search is invoked for a particular agent~$a_i$ that violates the node's constraints. Importantly, the search simultaneously explores agent positions in time and space, and while doing so ensures that the constraints~$C_N$ are satisfied. Once a path for the agent is found, the set of solution paths for $N$ is updated.

\subsection{Priority-Based Search}
Priority-based search (\pbs) is a recent approach for \mapf. It forgoes the completeness and optimality guarantees for the sake of computational efficiency. 
We now provide an overview of \pbs, and refer the reader to~\cite{ma2019searching} for the full description.
At its core, \pbs has some resemblance to \cbs, in the sense that it is a hierarchical approach with high and low level search. However, unlike \cbs which maintains space-time constraints between the agents in the high-level search, \pbs maintains priorities between agents.

On the high-level search, \pbs explores a \emph{priority tree} (PT), where a given node~$N$ of PT encodes a (partial) priority set $P_N=\{a_h \prec a_i,~a_j \prec a_l$,~\ldots \}. A priority $a_{i}\prec a_{j}$ means that agent $a_{i}$ has precedence over agent~$a_{j}$ whenever a low-level search is invoked (see below). In addition to the ordering, each PT node maintains single-agent paths that represent the current \mapf solution (possibly containing conflicts).
\pbs starts the high-level search with the tree root whose priority set is empty, and assigns to each agent its shortest path. Whenever \pbs expands a node $N$, it invokes a low-level search to compute a new set of paths which abide by the priority set $P_N$. If a collision between agents, e.g.,~$a_i$ and~$a_j$, is encountered in the new paths, \pbs generates two child PT nodes $N_1, N_2$ with the updated priority sets $P_{N_1}=P_N\cup \{a_i\prec a_j\}, P_{N_2}=P_N\cup \{a_j\prec a_i\}$, respectively. The high-level search chooses to expand at each iteration a PT node in a depth-first search manner. 
The high-level search terminates when a valid solution is a found at some node $N$, or when no more nodes for expansion remain, in which case, \pbs declares failure. 

The low-level search of \pbs proceeds in the following manner. For a given PT node $N$, \pbs performs a topological sort of the agents according to $P_N$ from high priority to low, and plans individual-agent paths based on the ordering. 
For a given topological ordering $(a'_{1},\ldots, a'_{k'})\subset \A$, for some $1\leq k'\leq k$, the low-level iterates over the~$k'$ agents in the topological ordering, and updates their paths such that they do not collide with any higher-priority agents. (Note that agents that do not appear on this list maintain their original plans.) It then checks whether collisions occur between all the agents combined.

\section{Algorithmic framework}
\label{sec:algs}
In this section we present our algorithmic contributions for tackling the \tmapf problem. We first describe a complete and optimal approach that is based on \cbs, and then proceed to a faster but incomplete \pbs-based method. 

\subsection{A \cbs-based approach for \tmapf}
We present an extension of the \cbs algorithm called \tfcbs, to solve the \tmapf problem. 
The main idea behind \tfcbs is to associate each mover agent $a^m_i\in \A^m$ with a particular movable obstacle $o^m_j\in \O^m$ and treat those two elements as one entity in both levels of the search.
In other words, $a^m_i$ can be thought of as agent that has two physical interpretations, that of the location of the actual agent, and the location of the obstacle $o^m_j$ it is required to move. 
The assignment of a mover to a movable obstacle is done once in the beginning of the execution of \tfcbs. In particular, we use a greedy heuristic where every mover agent $a^m_i \in \A^m$ is assigned to the closest movable obstacle $o^m_j\in \O^m$ according to the distance over $\G$, while ignoring obstacle locations, from $s^{ma}_i$ to $s^{mo}_j$. If an agent's closest obstacle is already taken by another agent, then the next-closest obstacle is assigned (and so on). We leave the study of stronger assignment approaches, such as the  Hungarian method~\cite{kuhn1955hungarian}, for future work. For the remainder of this section, we assume that each mover agent $a^m_i \in \A^m$ is associated with a specific movable obstacle that, w.l.o.g, is denote by $o^m_i$. 

Given that we treat $a^m_i$ and $o^m_i$ as one entity, we only impose in the high-level search of \tfcbs constraints on the agents of $\A^t$ and $\A^m$ but not on the movable obstacles. Next, we describe the constraints we use in \tfcbs, which generalize those of \cbs. We first consider constraints involving task agents. For a given $a^t_i\in \A^t$ the positive and negative constraints $(+,a^t_i,v,t)$ and $(-,a^t_i,v,t)$ are defined exactly as in the setting of  \cbs, i.e., $a^t_i$ should or should not visit $v$ at time $t$, respectively. 

Next, consider a mover agent $a^m_i$ and its assigned obstacle $o^m_i$. The positive constraint $(+,a^m_i,v,t)$ should be interpreted as the agent should visit $v$ at time $t$, with or without~$o^m_i$. 
The negative constraint $(-, a^m_i, v, t)$ simply means that~$a^m_i$ should not be in $v$ at time $t$, with or without $o^m_i$. 

Now  we introduce a new type of constraint to prevent collisions with movable obstacles before they are reached by the designated mover, which is not covered via the previous constraints. In particular, consider a movable obstacle $o^m_i$ and suppose that we have a lower bound $t$ on the time of arrival of agent $a^m_i$ to $s^{mo}_i$. In such a case we would like to inform task agents to avoid getting to $s^{mo}_i$ before time $t$, as they surely cannot reach this location since it is blocked and cannot be moved until after timestep $t$. Similarly, we would like to prevent mover agents that currently carry obstacles from reaching this vertex as well. Thus, we introduce the timed constraint $(s^{mo}_i,t)$ requiring that any agent reaching $s^{mo}_i$ before time $t$ is a mover agent that is not carrying an obstacle while passing through this vertex. 

Before wrapping up the description of the high-level search, we mention that \tfcbs maintains for every CT node its cost computed either using Eq.~\eqref{eq:without-mover} or Eq.~\eqref{eq:with-mover}. \tfcbs determines the next CT node to expand in a best-first search manner according to the node cost.

We now proceed to describe the low-level search of \tfcbs (visualized in Fig.~\ref{fig:mapf_intro}). For a task agent $a^t_i\in \A^t$ the search proceeds in a manner similar to \cbs using \algname{A*} and while adhering to the constraints of the current high-level node. Note that during this process the agent needs to avoid collisions with static obstacles, but collision avoidance  with movable obstacles is enforced through positive constraints over mover agents, e.g.,  $(+,a^m_i,v,\tau)$, or timed constraints of the form $(s^{mo}_i,\tau)$. 

When a low-level search for an agent $a^m_i$ is invoked, its search conceptually consists of two parts: 
(i)~$a^m_i$ moves from its start location $s^{ma}_i$ to the start location $s^{mo}_i$ of $o^m_i$, and 
(ii)~$a^m_i$ moves from $s^{mo}_i$ to $s^{mo}_i$, while carrying the obstacle~$o^m_j$. Note that in part (i) the agent~$a^m_i$ is free to visit vertices occupied by static or movable obstacles while abiding by the constraints, whereas in part (ii) the search only permits~$a^m_i$  to visit vertices that abide by the constraints and do not include obstacles. After the expansion of a node, new constraints are added to the child nodes, and their costs are updated, as in \cbs. In addition, a new lower bound $t$ for the arrival of a mover agent $a^i_m$ to its movable obstacle is computed, and its corresponding timed constraint is updated. 

\paragraph{Theoretical properties and computational complexity}
For a given assignment of mover agents to movable obstacles, \tfcbs derives the properties of \cbs (see~\cite{sharon2015conflict}) and is both complete and optimal (proof omitted).
Analyzing the running time is somewhat more complicated.
\citet{gordon2021revisiting} show that the number of CT node expansions for \cbs may be as high as $O\left((e|V|)^{|\A|C^*}\right)$, with $C^*$ being the cost of the optimal solution.
While our problem is not exactly the same, in our setting we need to account for the agents $\A^m$, which incur an exponential price in the worst case.
An exact analysis is left for future work.

\subsection{A \pbs-based approach for \tmapf}
\begin{figure}[tb]
     \centering
     \includegraphics[width=0.55\columnwidth]{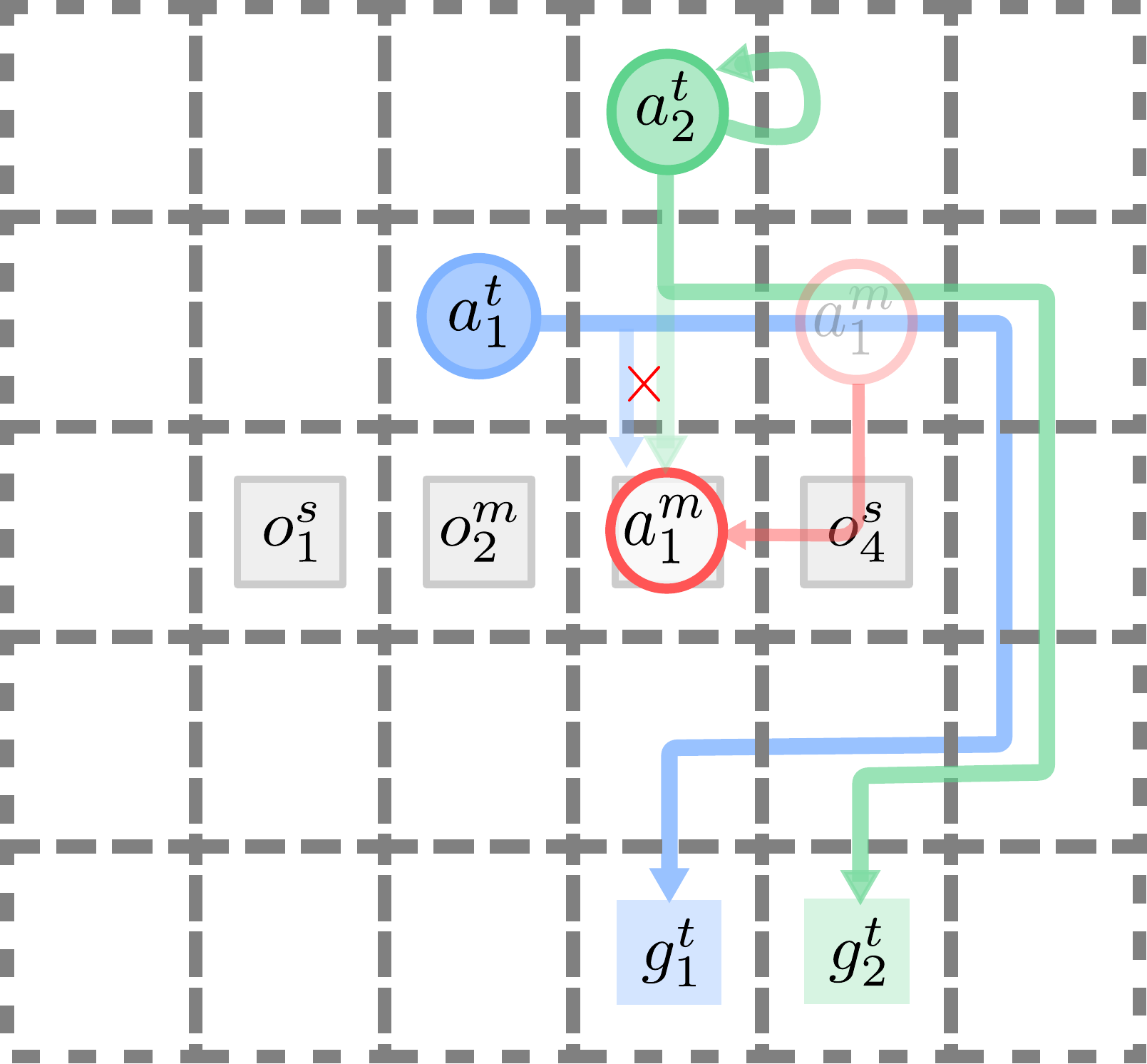}
    \caption{
        Illustration of the adverse side-effect of transitive ordering. Starting with a priority ordering $a^t_1 \prec a^t_2$ at a PT node, the low-level planner computes a path that causes agent~$a^t_1$ to collide with mover $a^m_1$ (at $s_3^{mo}$) while $a^t_2$ waits a single timestep.
        Since obstacle displacement cost by $a^m_1$ is greater than the path savings of $a^t_1$ (yet not accounting for the fact  that $a^t_2$ would also benefit from $a^m_1$ clearing the way), the collision between $a^t_1$ and $a^m_1$ would impose a priority $a^m_1 \prec a^t_1$. Then, transitive ordering implies a total ordering $\{a^m_1 \prec a^t_1 \prec a^t_2\}$ that causes both~$a^t_1$ and~$a^t_2$ to avoid colliding with the path of mover $a^m_1$, never utilizing the shortcut. 
        }
    \label{fig:pbs-problem}
\end{figure}
We describe a \pbs-based approach for \tmapf called \tfpbs. Similarly to \tfcbs, in \tfpbs we associate each mover agent $a^m_i\in \A^m$ with a particular movable obstacle $o^m_i\in \O^m$  and treat those two elements as one entity. On the high-level we explore a priority tree (PT) whose nodes encode priority sets between task and mover agents, as in \pbs. The low-level search for a given agent proceeds in a manner similar to \tfcbs in that for a task agent it searches for a path from its start to goal, and for a mover agent $a_i^m$ it searches for a path from its start location $s_i^{ma}$ to its obstacle's location $s_i^{mo}$, and finishes at~$s_i^{mo}$. 

We do, however, make a departure from the way priorities are treated in the high-level search within \pbs to accommodate the special circumstances of \tmapf: Recall that when solving the \mapf problem, \pbs enforces transitive ordering of priorities induced by a topological sort. For instance, given a priority set $\{a_3\prec a_1,~a_1\prec a_2\}$, for some node $N$ of the PT tree and some three agents $a_1,a_2,a_3\in \A^t \cup \A^m$, a topological sort would yield the total order $a_3\prec a_1 \prec a_2$, which implies that the low-level search would compute a path for $a_1$ while avoiding collision with $a_3$, and a path for~$a_2$ while avoiding both $a_3$ and $a_1$.

Such an approach can lead to poor utilization of the movers' capabilities of clearing shortcuts in \tmapf. Suppose that agents $a_1, a_2$ are task agents $a^t_1, a^t_2$ and $a_3$ is a mover $a^m_1$ whose shortest path in the low-level search is such that it would not move once it reaches its movable obstacle~$o^m_3$. Considering that $a^m_1\prec a^t_1$, agent $a^t_1$ would opt for a detour around a passage blocked by~$o^m_3$. Due to transitive ordering, $a^t_2$ would also avoid the blocked passage, even if going through it would significantly improve its path cost. This phenomenon is extended to the descendants of the current node, and could be exacerbated with a few more agents. In more general terms, $a^m_1$ chooses to block a shortcut, even though the shortcut can potentially serve multiple agents and the aggregate path cost savings across agents moving through the shortcut can offset the cost of obstacle displacement. 

In contrast, by reasoning about the exact priorities given in the priority set (rather than transitive ordering) we can avoid such a situation. For instance, if $a^t_2$ plans while avoiding collision only with $a^t_1$ (due to the constraint $a^t_1\prec a^t_2$) it may choose to go through the shortcut. This would cause a collision in the current PT node between $a^m_1$ and $a^t_2$, which could be resolved in offspring  PT nodes. 
 Fig.~\ref{fig:pbs-problem} depicts the problem with transitive ordering and illustrates the phenomenon from the previous paragraph.

Thus, within \tfpbs we employ a combined approach that utilizes direct prioritization whenever agents collide, and a greedy depth-first expansion scheme that steers the high-level search towards nodes with lower node cost in tie-breaks. 
Namely, the high-level search produces two child nodes and expands first the child with lower cost among the two. It continues exploring the descendants until a solution is found or until it reaches a node where the low-level search deems that an agent cannot reach its goal due to higher-ranking agents blocking the way.

\begin{figure}[tb]
     \centering
     \begin{subfigure}[b]{0.59\columnwidth}
         \centering
         \includegraphics[width=\columnwidth]{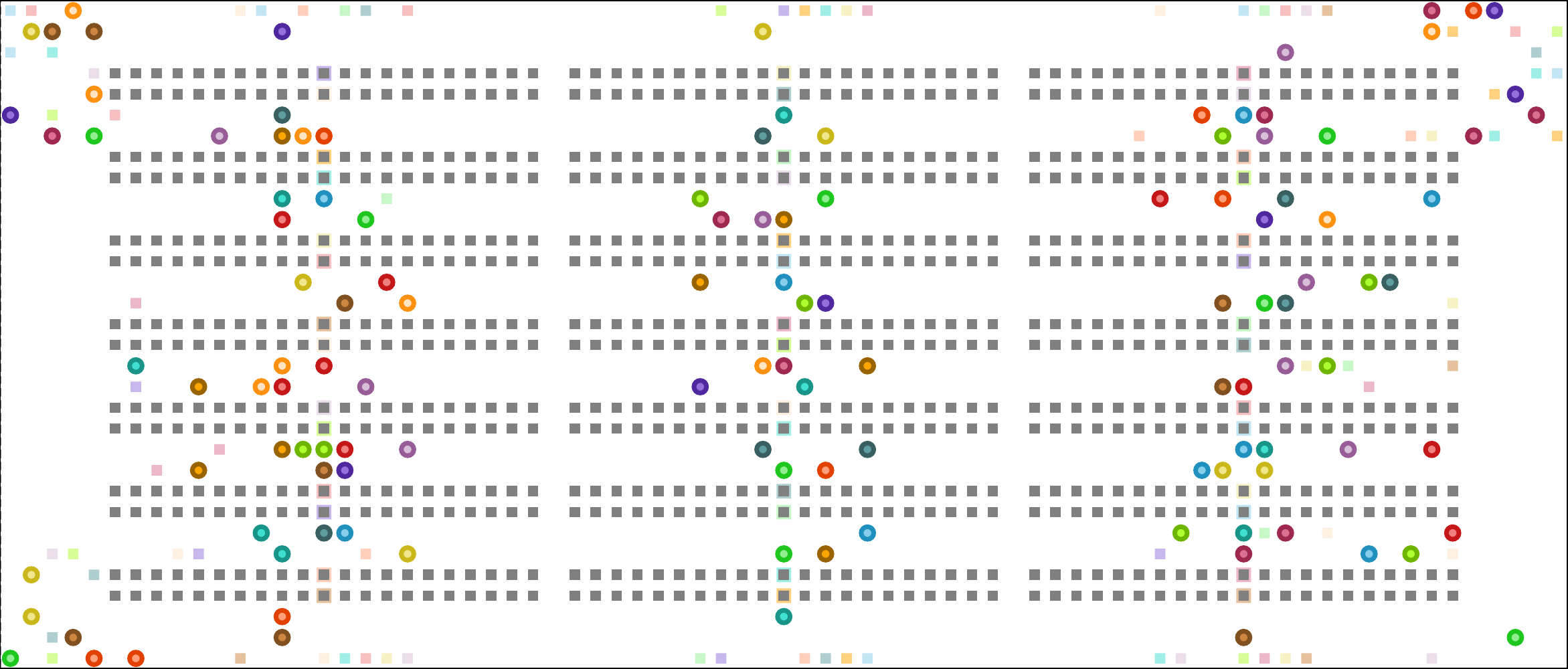}
         \caption{}
         \label{fig:eval-large}
     \end{subfigure}
     \hfill
     \begin{subfigure}[b]{0.39\columnwidth}
         \centering
         \includegraphics[width=\columnwidth]{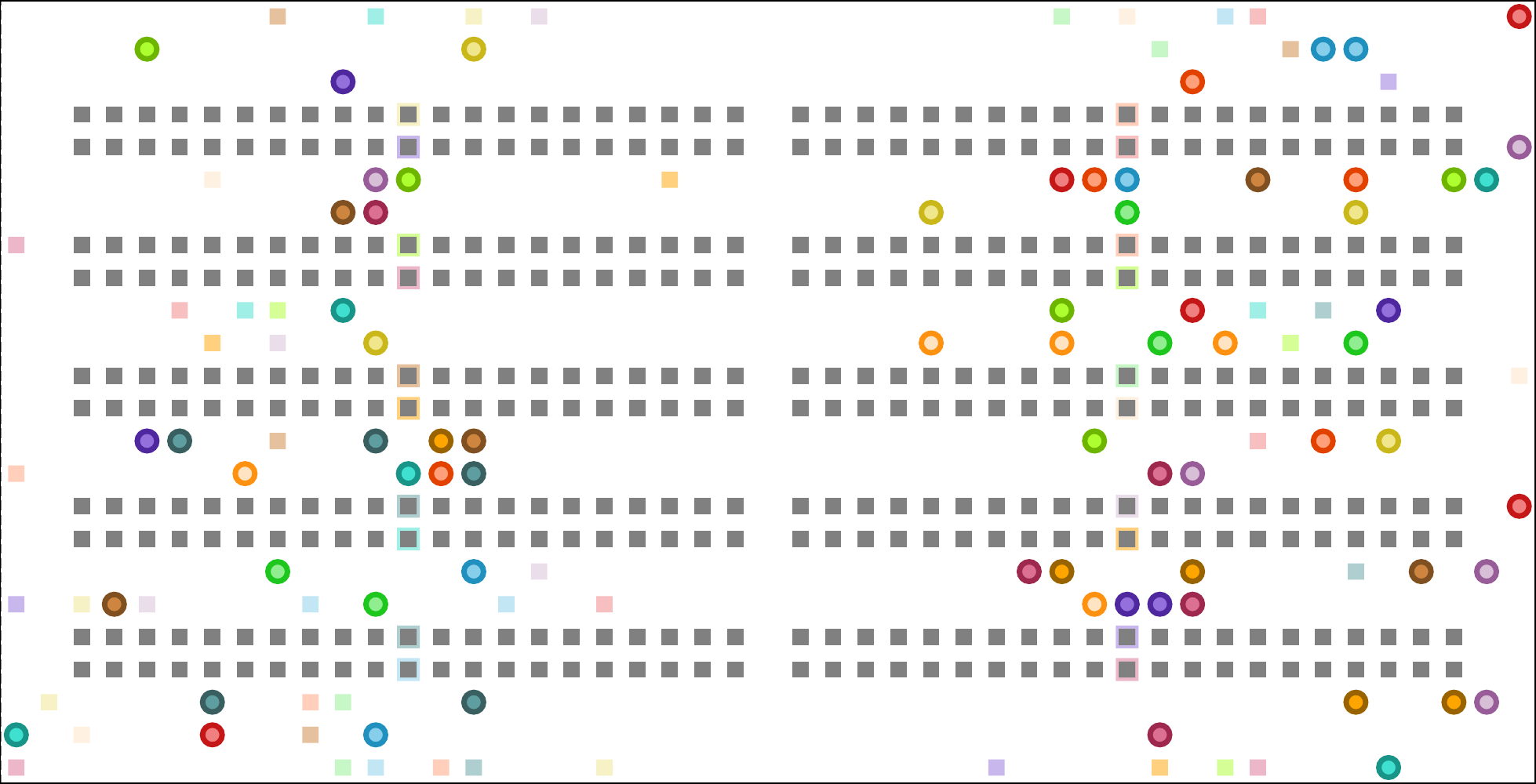}
         \caption{}
         \label{fig:eval-small}
     \end{subfigure}
    \caption{
        Warehouses used in our empirical evaluation.
        \protect{(\subref{fig:eval-large})} and 
        \protect{(\subref{fig:eval-small})} depict the
        \largeEnv and \smallEnv environments, 
        with~$80$ and $50$ task agents (colorful dots),
        respectively.
        Here rows of shelves (gray rectangles) form long narrow aisles
        and goals (colorful rectangles) are selected from nearby workstations around the periphery of each map.  
        %
        }
    \label{fig:eval-warehouse}
\end{figure}

\section{Evaluation}
\label{sec:eval}

A typical warehouse presents long rows of shelves that form narrow aisles, with workstations located around the perimeter of the map (as illustrated in Fig. \ref{fig:eval-warehouse}). In autonomous warehouses, longer aisles allow for greater storage capacity, but also lead to  constrained environments that quickly become congested as more agents are introduced. Hence, we evaluate our approach using maps inspired by autonomous warehouses with intentionally long aisles, and assess the impact of terraforming on measures of solution quality, node expansions, and success-rate. We conduct our experiments on two warehouse-like maps\footnote{Code and data will be made public  upon  publication.}: \\ %
\begin{itemize}
    \item \smallEnv \cite{felner2018adding,li2019disjoint} of size $24 \times 47$.
    \item \largeEnv \cite{li2020new} of size $32\times 75$.
\end{itemize}

For each map we vary the number of task-agents~$|A^t|$, and for every combination of map and $|A^t|$ we generate $10$ scenarios where the agents' start vertices are uniformly distributed. The agents' goal vertices are randomly selected to be either (1) around  workstations (at the perimeter of the map) or (2) empty vertices across the entire map (there is a 50/50 chance to choose from (1) or (2)). 
In this manner, we obtain a flow of agents both to and from workstations.

For \tmapf, recall that we make a simplifying assumption that our input also includes a set of movable obstacles~$\O^m$ as well as a set of mover agents $\A^m$, and require that $|\A^m| = |\O^m|$. Across all experiments, a set of $20$ and $42$ movable obstacles are selected for the \smallEnv and \largeEnv maps respectively, situated in the middle of every row of shelves. When comparing \tmapf with \mapf, the static environment treats all obstacles as static and solves strictly for task agents $\A^t$ (omitting mover agents $\A^m$).

We implemented the algorithms in Python and tested them on an Ubuntu machine with 4GB RAM and a 2.7GHz Intel~i7 CPU.

\begin{figure}[bt]
     \centering
    \includegraphics[width=\columnwidth]{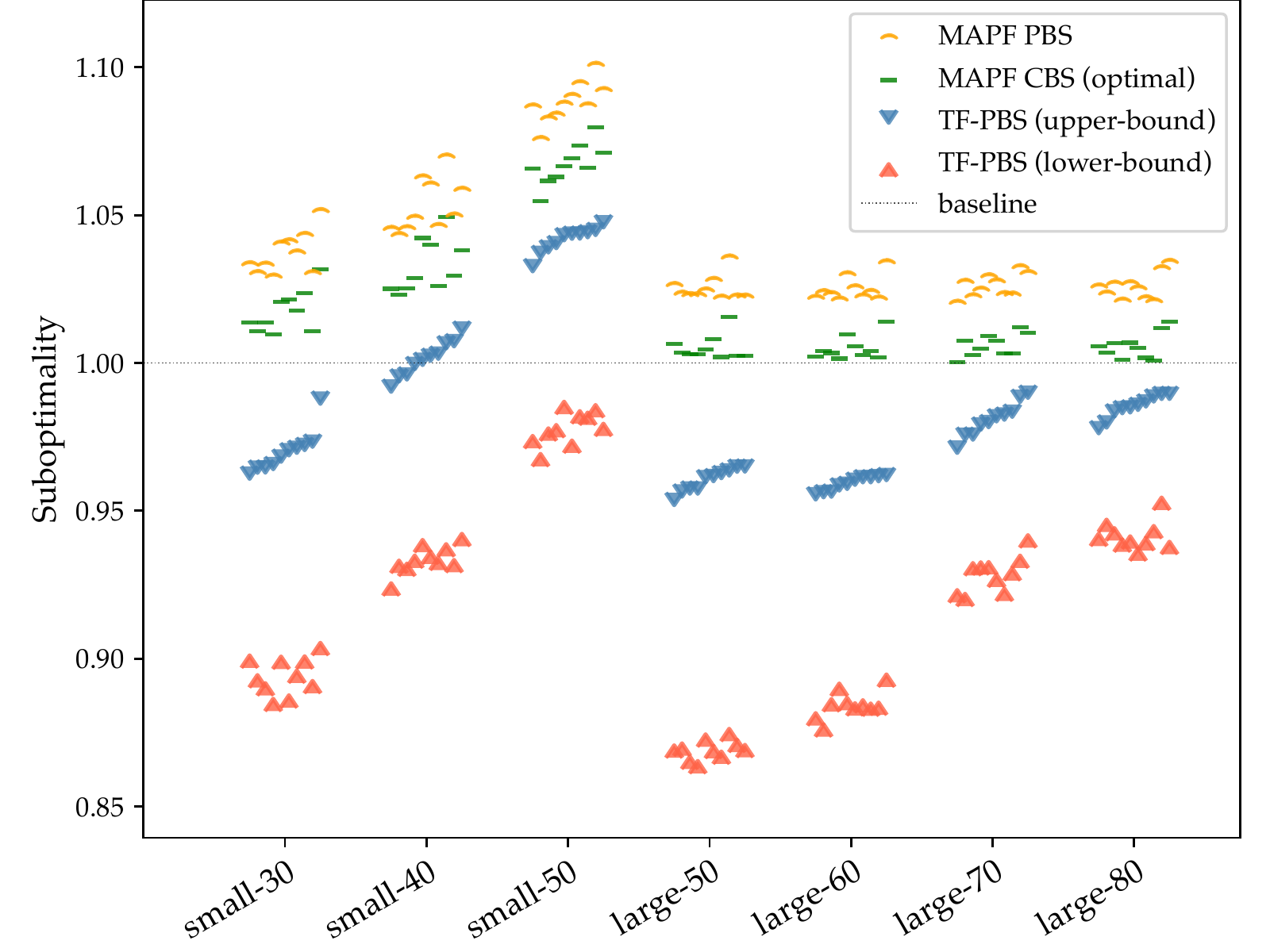}
    \caption{
        Solution quality comparison between \mapf-optimal solutions, \pbs solutions, and \tmapf solutions relative to the \emph{baseline} solution (dashed line). 
        }
    \label{fig:eval-soc}
\end{figure}

\paragraph{Solution quality.} We report solution quality as the \emph{sum of costs}, as shown in Figure \ref{fig:eval-soc}. The horizontal axis specifies the \smallEnv and \largeEnv warehouse maps, denoted \smallEnv-$n$ and \largeEnv-$n$, where $n=|\A^t|$. The vertical axis measures solution suboptimality relative to a lower-bound \mapf solution of ideal single-agent shortest paths, called \emph{baseline}. The baseline (dashed line) does not account for inter-agent collisions and interactions, which regularly degrade from the quality of the optimal solution. Fig. \ref{fig:eval-soc} shows the optimal \mapf solution (obtained with \cbs) as being above the baseline due to delays and congestion.

Next, we compare the solution cost obtained by \tfpbs for the \tmapf problem. 
Recall that here there are additional mover agents  ($20$ for \smallEnv and~$42$ for \largeEnv) to be considered by the search.
Fig.~\ref{fig:eval-soc} shows both the upper-bound cost (Eq.~\eqref{eq:with-mover}) and the lower-bound cost (Eq. \eqref{eq:without-mover}). 
Recall that the lower-bound reflects the path cost of regular agents and of mover agents as they carry their assigned obstacle, whereas the upper-bound also accounts for movers' path cost en-route to their obstacle. The results suggest that terraforming has the capacity to \emph{outperform} the optimal solution available for a static environment by alleviating bottlenecks and long detours. Interestingly, the figure also demonstrates the potential to even surpass the baseline of a given scenario, thanks to the ability of mover agents in \tmapf to create shortcuts. 

\begin{figure}[bt]
     \centering
    \includegraphics[width=\columnwidth]{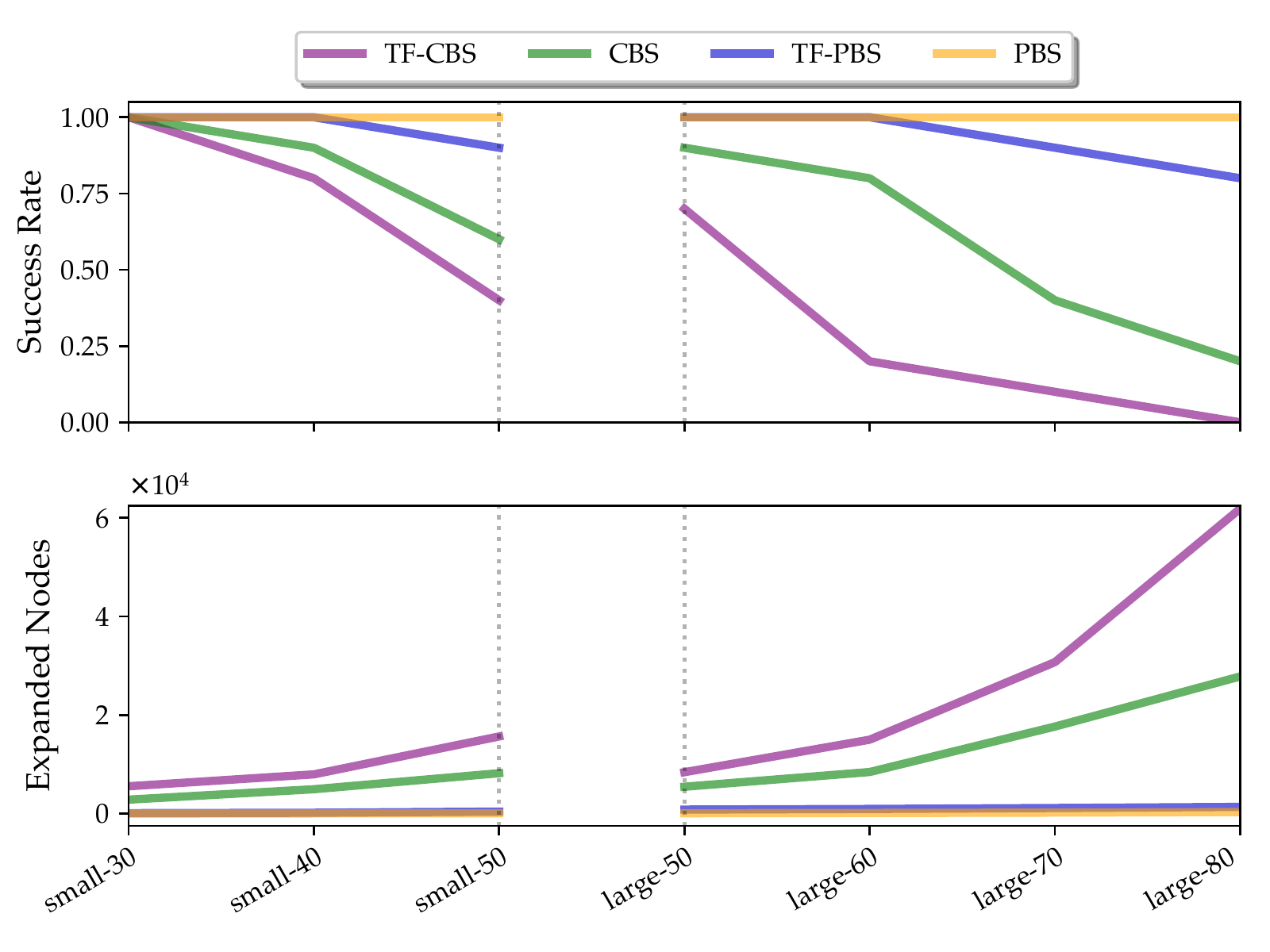}
    \caption{
        Comparison of success rate (top) and expanded nodes (bottom) between \tfcbs, \cbs, \tfpbs and \pbs. }
    \label{fig:eval-warehouse-success-expanded}
\end{figure}

\paragraph{Success rate and node expansions.}
Success rate is measured by the number of scenarios solved within a $5$ minute timeout, and use the number of (high-level) expanded nodes as a proxy for average runtime necessary for each approach to obtain a solution. Figure \ref{fig:eval-warehouse-success-expanded} shows the success rate and expanded node count for the various combinations of maps and number of task agents~$|\A^t|$. We re-iterate that for \tmapf, additional mover agents are introduced ($20$ for \smallEnv and~$42$ for \largeEnv). As a result, the total agent count~$|\A|$ for \tmapf is greater, which makes it more computationally challenging to solve than \mapf. As expected from a congested warehouse environment, we see a rapid deterioration in success rate as more agents are introduced. This is especially pronounced in \tfcbs than \cbs. In terms of the number of high-level node expansions, we see a steep rise in \cbs, and to a greater extent, in \tfcbs.

An encouraging trend is evident with \tfpbs, which is more robust in terms of success rate and node expansions relative to the other approaches. The warehouse environment presents narrow corridors that are prone to head-on collisions between two or more agents. This is where the adaptive agent-priority assignment of \tfpbs has a demonstrable advantage. When two agents collide, \tfpbs imposes a priority ordering that prevents future collisions between them. In this manner, local collisions between agents elicit lasting constraints on the search space that partially eliminate unnecessary explorations of the Priority Tree (PT), thus empirically reducing the number of node expansions necessary to reach a solution.

\ignore{
\begin{table}[t]
\centering
\begin{tabular}{|c|c|c|c|c|c|c|}
\multicolumn{2}{c|}{~} & \multicolumn{2}{c|}{ \emph{flowtime}} & \multicolumn{2}{c|}{\emph{latency}} \\ 
\hline
 \multicolumn{1}{|c}{~} & \multicolumn{1}{c|}{\small{agents}} & \tiny{\tfcbs} & \tiny{\tfpbs} & \tiny{\tfcbs} & \tiny{\tfpbs} \\ 
\hline
\multirow{3}{*}{\rotate{small}} & 10 & $1.00$ & $1.00$ & $1.00$ & $1.00$ \\
& 20 & $1.00$ & $1.00$ & $1.00$ & $1.00$ \\
& 30 & $0.95$ & $0.95$ & $0.95$ & $0.95$ \\ 
\hline
\multirow{3}{*}{\rotate{large}} & 50 & $1.00$ & $1.00$ & $1.00$ & $1.00$ \\
& 60 & $1.00$ & $1.00$ & $1.00$ & $1.00$ \\
& 70 & $0.95$ & $0.95$ & $0.95$ & $0.95$ \\ 
\hline
\end{tabular}
\caption{Performance \tfcbs and \tfpbs compared to \cbs. \textcolor{red}{TODO:Placeholder - not actual results}}
\label{table:comparison}
\end{table}
}

\section{Discussion and Future Work}
\label{sec:discussion}
In this work we explored the potential that terraforming has for \mapf-like problems.
As demonstrated in our evaluation, the result is a form of emergent collaboration,
in which agents create shortcuts and reduce the overall cost beyond what can be achieved in the same static environment.

To harness the full potential of \tmapf, we envision its application to \lmapf where agents en-route to collect an item can serve as mover agents, hence moving one or more movable objects with minimal overhead. 
This blurs the simplifying assumption we made in our problem formulation where  there is a clear distinction between mover and task agents, and motivates the cost functions  we introduced for \tmapf.  This also requires lifting assumptions \textbf{A1} and~\textbf{A2}.

The main challenge we see is how to dynamically assign movable obstacles to agents that are not currently carrying items.
We consider two alternatives:
In the first alternative, agents not carrying items plan to their original goal and are assigned movable obstacles that lie on or near their already-planned path. This will probably incur little overhead for the agent but may result in an assignment to a movable obstacle whose displacement offers only a minor benefit.
In the second alternative, task agents that identify that moving a movable obstacle will dramatically reduce their path length will ``request'' that the obstacle be moved. This will trigger an assignment of that specific obstacle to the closest agent not carrying an item. This can be seen as the complementary case to the first alternative where moving an obstacle may incur a large overhead for the agent but results in an assignment to a movable obstacle that can significantly improve costs if moved.

\begin{figure}[bt]
    \centering
    \includegraphics[width=0.75\columnwidth]{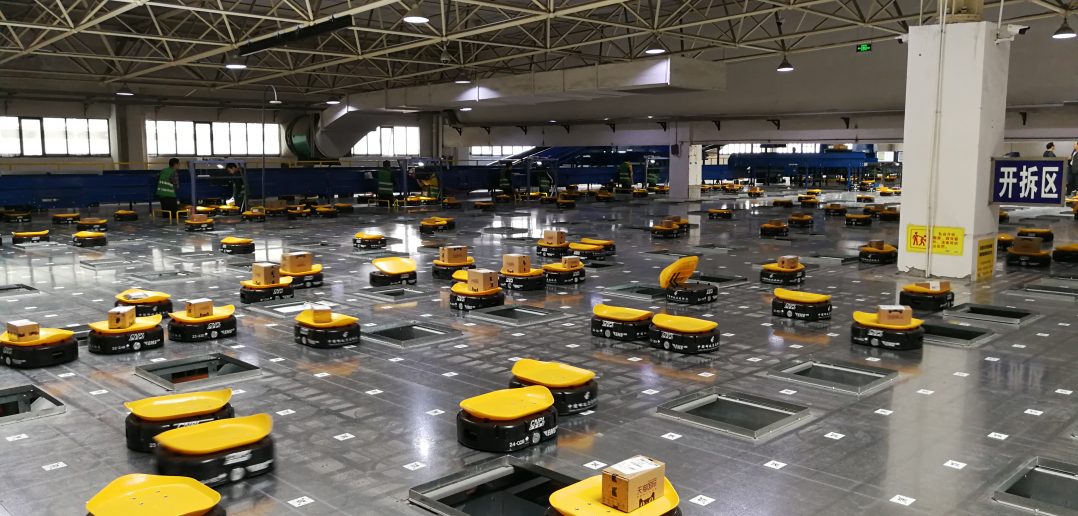}
    \caption{
        Sortation center. Figure adapted from \url{https://tinyurl.com/bdj8n8kx}.}
    \label{fig:sortation}
\end{figure}

Finally, we foresee applications of our work beyond our motivating example of \mapf in warehouses.
For example, consider sortation centers (see~\cite{KouP0KK20} and visualization in Fig.~\ref{fig:sortation}) where agents need to a reach a dropoff station, obtain a parcel for delivery, and then deliver the parcel to a sorting bin. 
We envision the sorting bins as having a mechanism that can be used to automatically cover them (possibly incurring time). This will allow agents to pass over covered bins hence reducing their path length. In the context of our work, these bins are the movable obstacles considered in this work. There are several challenges that need to be addressed (both mechanical as well as algorithmic), but this showcases the potential power of terraforming for increasing throughput of automated logistic centers.

\bibliography{aaai22.bib}

\end{document}